\documentclass[a4paper,fleqn]{cas-dc}

\usepackage[numbers]{natbib}
\usepackage{float}
\usepackage{graphicx}
\usepackage{amsmath,amssymb,amsfonts}
\usepackage{hyperref}
\usepackage[ruled,vlined]{algorithm2e}
\usepackage{multirow}
\usepackage{tabularx, booktabs}
\usepackage{subfig}
\usepackage{mathtools}
\usepackage{arydshln}
\usepackage{xcolor}
\usepackage{mwe}
\usepackage{xcolor}

\definecolor{red}{rgb}{1.00,0.00,0.00}
\definecolor{blue}{rgb}{0.00,0.00,1.00}

\usepackage{amsmath}
\DeclareMathOperator*{\argmax}{argmax}   
\usepackage{flushend} 

\def\tsc#1{\csdef{#1}{\textsc{\lowercase{#1}}\xspace}}
\tsc{WGM}
\tsc{QE}
\tsc{EP}
\tsc{PMS}
\tsc{BEC}
\tsc{DE}

\begin{document}

\let\WriteBookmarks\relax
\def\floatpagepagefraction{1}
\def\textpagefraction{.001}

\shorttitle{MORE: Simultaneous Multi-View 3D Object Recognition and Pose Estimation}
\shortauthors{T. Parisotto et~al.}

\title [mode = title]{MORE: Simultaneous Multi-View 3D Object Recognition and Pose Estimation}


\author[1]{Tommaso Parisotto,}
\address[1]{Department of Artificial Intelligence, Bernoulli Institute, University of Groningen, The Netherlands.}

\author[2]{Subhaditya Mukherjee,}
\address[2]{Department of Artificial Intelligence, Bernoulli Institute, University of Groningen, The Netherlands.}

\author[3]{Hamidreza Kasaei}
\cormark[1]
\fnmark[1]
\ead{hamidreza.kasaei@rug.nl}
\ead[url]{https://www.ai.rug.nl/irl-lab/}
\address[3]{Department of Artificial Intelligence, Bernoulli Institute, University of Groningen, The Netherlands.}


\begin{abstract}
Simultaneous object recognition and pose estimation are two key functionalities for robots to safely interact with humans as well as environments.  Although both object recognition and pose estimation use visual input, most state-of-the-art tackles them as two separate problems since the former needs a view-invariant representation while object pose estimation necessitates a view-dependent description. Nowadays, multi-view Convolutional Neural Network (MVCNN) approaches show state-of-the-art classification performance. Although MVCNN object recognition has been widely explored, there has been very little research on multi-view object pose estimation methods, and even less on addressing these two problems simultaneously. The pose of virtual cameras in MVCNN methods is often pre-defined in advance, leading to bound the application of such approaches. In this paper, we propose an approach capable of handling object recognition and pose estimation simultaneously. In particular, we develop a deep object-agnostic entropy estimation model, capable of predicting the best viewpoints of a given 3D object. The obtained views of the object are then fed to the network to simultaneously predict the pose and category label of the target object. Experimental results showed that the views obtained from such positions are descriptive enough to achieve a good accuracy score. Furthermore, we designed a real-life serve drink scenario to demonstrate how well the proposed approach worked in real robot tasks. We will release code, data upon acceptance.
\end{abstract}


\maketitle


\section{Introduction}

\begin{figure}[!h]
    \centering
    \includegraphics[width=\linewidth, trim={0 0cm 0 0cm}, clip]{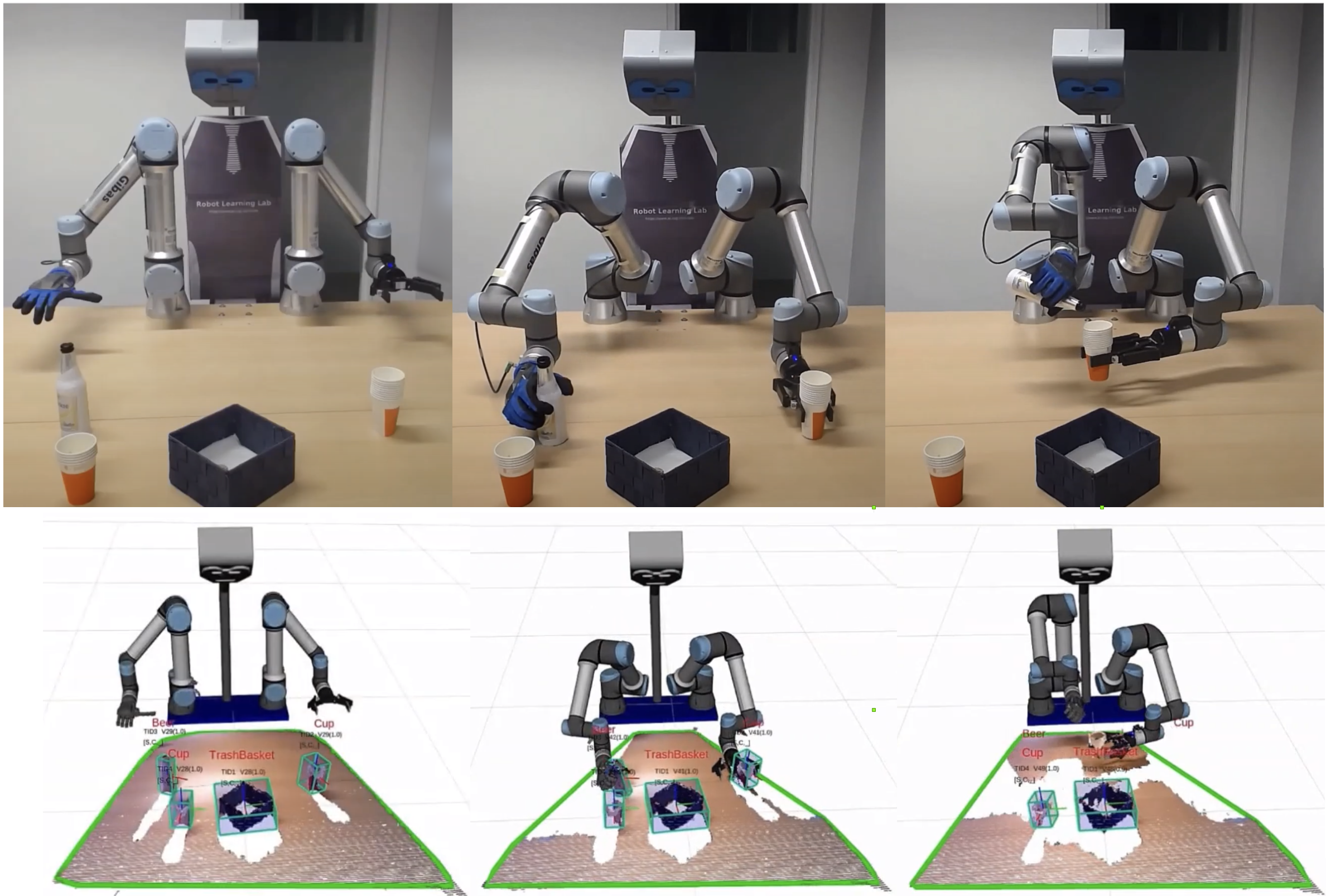}
    \caption{In this example, the robot is instructed to perform the serve drink task. To achieve this task successfully, the robot needs to know which objects exist in the scene, and where they are. Then, it should plan a trajectory for grasping the bottle object with one arm and one of the cups with the other arm. Finally, it serves the drink by moving the beer on top of the cup and pouring it into the cup.}
    \label{fig:example}
    \vspace{-2mm}
\end{figure}

Nowadays, robots are leaving pre-defined setting and helping humans in many collaborative tasks in both industrial and human-centric environments. In order to safely interact with users and environments, robots need to recognize a range of objects and estimate their poses precisely from different perspectives. It is a challenging task due to high demand for accurate object recognition and precise pose estimation, as the output of these tasks {are} used as input for the purpose of object manipulation. For instance, consider the task of serving beer as shown in Fig.~\ref{fig:example}. To accomplish this task successfully, the robot first needs to know which objects exist in the scene and where they are (Fig.~\ref{fig:example} \textit{first column}). Then, it should plan a trajectory to grasp the beer and one of the cups (Fig.~\ref{fig:example} \textit{middle column}), and finally, move the beer on top of the cup, and pour the beer into the cup (Fig.~\ref{fig:example} \textit{last column}). Although object recognition and pose estimation tasks both require visual information as input, they are often contradicting from a problem definition point of view. In particular, a robot needs to learn pose-invariant features of objects to be able to recognize them accurately from different viewpoints; 
\cite{su2015multi}In contrast, the robot requires to learn pose-dependent features of objects to be able to estimate their pose. This is the main reason that state-of-the-art approaches address object recognition and pose estimation often as two separate problems. Recent multi-view deep learning approaches achieved the best results in both object recognition and pose estimation when they tackled these problems separately. The pose of virtual cameras in MVCNN methods is defined in advance, leading to limitations in the application of such approaches.

\begin{figure*}[!t]
    \centering
    \includegraphics[width=\linewidth, trim={0 0cm 0 0cm}, clip]{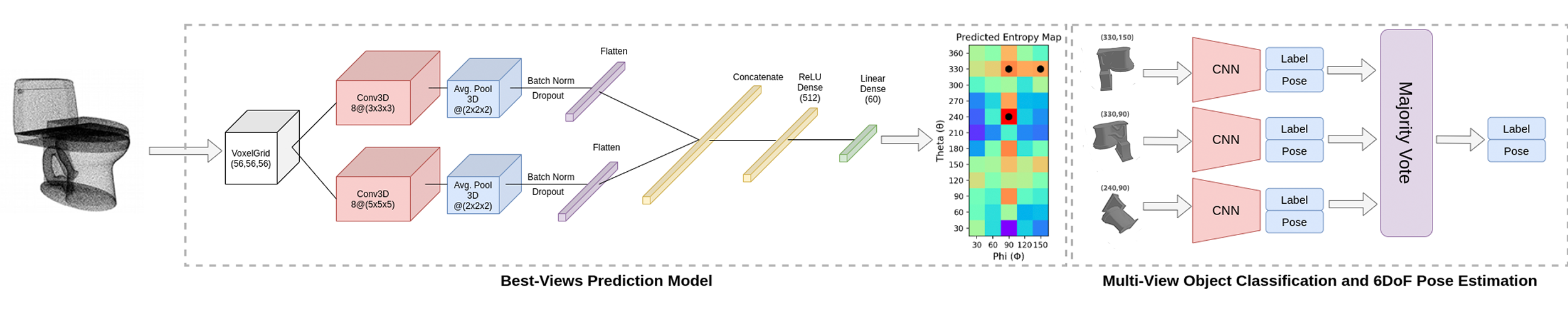}
    \vspace{-5mm}
    \caption{The 3D object is transformed into a $56\times56\times56$ voxel grid which is the input size for {MORE}. The entropy model uses a branching architecture that converges into a fully connected layer. The output of the entropy model is a 60-values vector which is reshaped into a 2D entropy map. From the entropy map a peak selection algorithm returns the coordinates of the local maxima. We extract the views corresponding to those coordinates and we use them as input for the CNNs. The label and pose predictions are pooled by majority vote, resulting in the final prediction. 
    }
    \label{fig:overview}
\end{figure*}

There are still many challenges to overcome, even though many problems have already been understood and solved successfully. Simultaneous object recognition and pose estimation is one of the challenges that require more research. In this paper, we take a step towards addressing this issue in the context of service robotics, by proposing an approach to handle object recognition and pose estimation simultaneously by sharing representations between these tasks. This is done by an object-agnostic entropy estimation model that automatically predicts the most informative viewpoints of a given 3D object directly and more efficiently. These predicted views are then fed into a viewpoint network to simultaneously predict the pose and category label of the target object as shown in Fig.~\ref{fig:overview}. {By using a single model for both tasks, MORE reduces the overall complexity and memory requirements of a robotics pipeline where real-time responsiveness is crucial. We also performed real-robot experiments to show the usefulness of the proposed approach in real-world scenarios.} The main contributions of this work are as follows:

\begin{itemize}
    \item We present a novel approach to simultaneously recognize objects and estimate their poses using a framework that first predicts the best views and then uses it for both tasks at once.
    \item We extensively evaluate the proposed approach and on publicly available dataset and achieved state-of-the-art object recognition accuracy of 98.26\% and 96.52\% on ModelNet10 and ModelNet40 datasets, respectively. 
    \item To the benefit of research communities, we release the source code and the trained models, making it possible to reproduce our results.
\end{itemize}

\section{Related Work}

Three-dimensional (3D) object recognition and pose estimation have been under investigation for a long time in both computer vision and robotics communities. Although an exhaustive survey of recent deep learning based approaches is beyond the scope of this paper, we {review} the main efforts.

There are substantially three main approaches for CNN-based object recognition: volume-based, point-based, and view-based approaches. Volume-based approaches use volumetric representation of data, in particular they employ voxelization methods to obtain a uniform representation for all input objects. The obtained representation is then used as input of the network. 
Point-based approaches are popular with data retrieved with 2.5D depth sensors. These sensors capture a dense set of depth samples from the scene, representing the surface of the objects as a collection of points in the Euclidean space. Point-based neural networks learn features about the positional relations between points on the surfaces of objects. View-based approaches use one or more images representations of the objects, usually captured with a camera from a specific viewpoint. CNNs trained on such representations learn features from the visible attributes of the objects. The availability of 3D data usually induces to directly apply recognition algorithms on 3D data, however, it has been shown that view-based methods outperformed other methods and achieved better performance.

Qi et al.~\cite{qi2016volumetric} provides a comprehensive study on voxel-based CNNs and multi-view CNNs for three dimensional object classification, stating that empirical results from the view-based and volume-based types of CNNs exhibit a large gap, indicating that existing volumetric CNN architectures and approaches are unable to fully exploit the power of 3D representations. Among voxel-based systems, one of the earliest works would be 3D ShapeNets~\cite{wu20153d} which developed a Convolutional Deep Belief Network to learn probability distributions of binary occupancy grids. VoxNet~\cite{voxnet} was designed to tackle object recognition by integrating the voxel representation to deal with large amounts of point cloud data. FPNN~\cite{fpnn} employed field probing filters to efficiently extract features from voxel data. 3D-GAN~\cite{3dgan} implemented Generative Adversarial Networks (GAN) to generate 3D objects from a probabilistic space and obtain a object descriptor from an adversarial discriminator. DensePoint~\cite{liu2019densepoint} was proposed as a variant grid CNN to find local patterns and learn useful hierarchies from point clouds. This makes the network particularly good at object identification. Kumawat et al., proposed a variant of 3D Conv layers, the ReLPV block, which applies multiple local STFTs to neighboring points and linearly combines the output~\cite{kumawat2019lp}. The LP-3DCNN was created using such blocks and was shown to be less memory intensive by having lesser trainable parameters than a regular 3D convolution. A variant of the standard convolution was introduced by Liu et al.~\cite{liu2019relationshape} , where the geometric relations of points in their respective cloud was used to create the RS-CNN. This operation was intended to better model spatial layout that would in turn make it better at understanding shapes from the point cloud.

Recent approaches showed that it is possible to achieve significant improvements in classification accuracy by using collections of rendered views of 3D objects~\cite{su2015multi,kanezaki_rotationnet_2021}. In particular, Su et al.,~\cite{su2015multi} obtained object's views by retrieving 2D projections of the object with a set of virtual cameras positioned in a regular setup. The authors opted for a fixed number of virtual camera points, positioning of the cameras on a regular structure around the objects. They demonstrated that a convolutional neural network, trained on a fixed set of rendered views of a 3D shape, could outperform most architectures trained on three-dimensional structured data.  It was shown that in many cases a single view already achieves satisfying classification accuracy. Jiang et al. introduced MLVCNN \cite{jiang2019mlvcnn}, which used multiple loops of views to extract hierarchical relationships between multiple views. The obtained representations were used to generate descriptive descriptions for 3D object classification. Zhang et al. tackled some of the faults of multiple hyper-graph transductive learning and propose iMHL~\cite{zhang2018inductive} where the hypergraph embedding is learnt offline while the test samples are classified in an online manner. This inductive approach suffers from not being able to rely on existing projection matrices if new classes are added to the data but is much faster than performing the learning transductively.

OrthographicNet \cite{kasaei_orthographicnet_2020} was designed for a more general online learning situation where the model was required to learn new classes after the initial training, with very few examples. This thus required the network to learn to model more rotation and scale in-variance and improved performance of learning new objects. Orthographic projection was used along with MobileNetV2 \cite{sandler2018mobilenetv2} to create a novel network that would, in collaboration with a simulated teacher - student approach, learn and update new features in an online manner. 

Kanezaki et al. proposed a multi-view CNN based approach namely RotationNet~\cite{kanezaki_rotationnet_2021}, which achieved SOTA. They proposed a CNN-based model which takes multi-views images as input and jointly estimates its pose and object category. Viewpoint labels are learned in an unsupervised manner during the training and the architecture is designed to use only a partial set of views for inference. Unlike MVCNNs, their method is able to classify an object using a partial set of images that may be sequentially observed by a moving camera. The system infers the probability of a retrieved view to match the camera position it has been taken from, subsequently determining the orientation of the object.

Another type of modeling multiple view data was proposed by Khan et al. \cite{khan_unsupervised_2019} where the authors worked on a finer 3D representation generation. A primitive discovery making use of physical properties and modeled by a higher order CRF was shown. The model also learns to differentiate between changes in the view and shape. This type of modelling leads to a more compressed representation as compared to voxels and similar data structures. It was also acknowledged that due to some of the objects being hollow, missing voxels led to a deterioration of accuracy.

Joint learning of object classification and pose estimation has already been unraveled by several researches~\cite{ma_learning_2019, kanezaki_rotationnet_2021, xuan_mv-c3d_2019, zhang_inductive_2018}
, however, very few of them address inter-class feature learning for pose alignment. 
Ma et al. \cite{ma_learning_2019} showed that a combination of CNNs and LSTMs could be used to create a more robust descriptor of the object by {taking into account multiple low level features} and then performing voting in order to improve performance. They also used a highway network layer to further ensure less loss of information between the two different types of networks. 
Xuan et al. \cite{xuan_mv-c3d_2019} took a similar route and proposed a 3D CNN that could take the multiple view points into account and better understand the object. This would work even if the object was only partially visible. The set of views of an object are taken fully as the input in order to learn the spatial relations between the views. The caveat being, if there were not enough views present, the model could not learn an accurate representation.

It has been proved beneficial to share appearance information across classes to simultaneously solve for object classification and pose estimation~\cite{kuznetsova2016exploiting}. Elhoseiny et al. \cite{elhoseiny2016comparative} studied CNNs for joint object classification and pose estimation based on multi-view images, discussing architectures of the following archetypes: Parallel Model (PM) consisting of two base networks running in parallel; Cross-Product Model (CPM) explores a way to combine categorization and pose estimation by building a last layer capable of capturing both; Late Branching Model (LBM) splits the network into two last layers, each designed to be specific to the two tasks; Early Branching Model (EBM) is  similar to LBM, however the branching is moved to an earlier layer in the network. 
While their method takes a single image as input for its prediction, later works focused on how to aggregate predictions from multiple images captured from different viewpoints \cite{kanezaki_rotationnet_2021}.

The best-view selection corresponds to the automated task of selecting the most representative view of a 3D model. Dutagaci et al. \cite{dutagaci2010benchmark} provides a benchmark for the evaluation of best-view algorithms and a survey on popular methods of best-view selection. 
The algorithms discussed by Dutagaci et al. differ with respect to the descriptor they use to assess the goodness of a view, which are assumed to measure the geometric complexity of the visible surface of an object. This survey was further continued by \cite{bonaventura_survey_2018} where the authors tried to unify the approaches from the previous paper and newer ones that were created in recent years. They proposed a new framework to do the same.

\section{{Proposed Methodology}}
We propose a deep learning approach to infer the best-views of a 3D model. The obtained views are then used to perform both object recognition and pose estimation tasks. We subdivide the problem in two main tasks, the first being the best-view prediction and the second being the multi-view based object recognition and pose estimation.

\subsection{Best-views prediction model}

The main objective is to design a model that predicts which point of views are most informative. In this vein, we first need to define how we measure quantitatively the goodness of an object view. From the information theory, we can calculate the expected information gain from various metrics (variance, entropy, etc.). Among these metrics, viewpoint entropy is a good proxy for expected information gain~\cite{thrun2002probabilistic}. In particular, by choosing views that cover more of the object's surface, the likelihood of estimating the pose and identifying the label of the object {increases. This is because viewpoints that capture high entropy areas tend to provide more informative data compared to
those that capture low entropy areas. In MORE, the view ranking procedure is based on viewpoint
entropy, which takes into account both the number of occupied pixels and their values.}
Therefore, we evaluate the quantity of information for each view by calculating the entropy of depth image captured from the same viewpoint with the definition from {Shannon's information theory \cite{shannon2001mathematical}}:
\begin{equation} \label{shannon}
    H(X) = -\sum^{n}_{i=1}p(x_{i})\log{p(x_{i})}
\end{equation}

\noindent where $x_i$ represents the value of $i$th pixel~\cite{kasaei2018perceiving}\cite{Sock_2017_ICCV}. 
Although it is possible to render several views of an object and then select the best views by computing the view entropy for each view, such approaches are computationally expensive and not appropriate for robotic applications. Our intention is to replace multiple computational steps with a single neural network inference, thus improving efficiency. This training needs to be done once to learn an object-agnostic best-view prediction function. {It is worth mentioning that the view selection function can be customized to suit the criteria of any other task with ease.
The proposed entropy calculation is not rely on the size of the object. Although considering size might lead to a slightly better performance in differentiating similar objects, doing so would increase computation time and is not considered.}

Towards this goal, instead of a multi-label classification problem where we classified each viewpoint as informative or not, we defined the problem as a regression to infer the entropy values of every viewpoint, generating a spherical entropy map of the object. The entropy map $H(\phi,\theta)$ is learned in the form of a 2D function that maps two spherical coordinates, $\phi$ and $\theta$ necessary to identify the viewpoints on a sphere around the object to the inferred entropy values: $H : (\phi,~\theta) \xrightarrow{} h$. {Since MORE can generate the entropy map from any angle, the initial angle does not influence the results.} {For a view $v$}, the coordinates of the most informative views are then obtained by evaluating the peaks of the entropy map: 

\begin{equation}
\{ (\phi_v, \theta_v) \}= \argmax_{\phi, \theta}(\frac{d^2H}{d\phi d\theta} = 0)
\end{equation}


We design a CNN approach to estimate an entropy map for a given object. As shown in Fig.~\ref{fig:overview}, we employ two convolutional branches with kernels of different sizes separating the flow of the graph from the input layers. Supposedly the different kernel size help the network identifying high level features of different scales. The output of the convolutional branches both receive average pooling, batch normalization and dropout before being transformed into flat vectors. The outputs of the branches are then concatenated in a single vector and sent as input to a fully connected hidden layer. The last layer is a fully connected output layer with linear activation that outputs 60 entropy values (see Fig.~\ref{fig:overview}). Note that a linear activation is used instead of one like softmax, because the outputs are entropy values and not probabilities. It is to be noted that a softmax activation can also be used if the view values are normalized. To make sure the activations were proportional to the input, a linear activation was used here. The optimal number of filters for the convolutional layers and the number of units in the hidden layer was estimated empirically with a hyper-parameters based on the search algorithm Hyperband~\cite{li2017hyperband}, a bandit-based approach to hyper-parameters optimization that speeds up random search through adaptive resource allocation and early-stopping. It evaluates architectures by training a set of configurations for a limited number of epochs and carrying the evaluation only for the most promising half until it reaches the best set of parameters. We used the Adam optimizer with dynamical learning rate starting at $5e-5$ with a reduction on the plateau of factor 0.3 and Mean Absolute Error as a loss function. To improve training speed, Mixed precision training \cite{micikevicius2017mixed} was also used where lower precision values are used in some parts of the network to reduce computational effort.

Towards this end, the first step is to generate a dataset taking depth images from several views of a collection of objects, in particular we took images from $60$ positions, regularly distributed on a sphere. The virtual cameras are positioned on $12$ points on a section ring of the sphere, each one at an angle of $30$ degrees from the next one. The sphere is circled by $5$ rings, parallels to the horizontal axis of the object, which are looking at the center of the sphere from each at an angle of $30$ degrees from the next one, cutting the sphere at $30$, $60$, $90$, $120$ and $150$ degrees from the vertical axis of the object. The structure of the camera positions ensures we obtain a complete overview of any object while having a limited number of fixed positions. Once we have the positions for the cameras, we take a grayscale depth image of $224\times224$ pixels of the object for each of the $60$ views. We can then evaluate the quantity of information in each view by calculating their entropy (details are presented in Section~\ref{dataset}).

We initially trained the model as a best-view classifier, however, such configuration tended to provide a list of best-views based on high entropy values instead of learning the relationship between silhouettes and entropy. {Since the ModelNet dataset is an imbalanced dataset, such approaches do not work well.} We opted for a more efficient solution: we built the dataset by matching each 3D object to its entropy values and trained the 3D-CNN to infer by regression the values from any 3D model. With this solution we observed the network generalizes better on new data and it allows for a more precise evaluation of the best-views.

\begin{figure}
    \centering
    \includegraphics[width=\linewidth,trim={3cm 0.5cm 3cm 0}, clip]{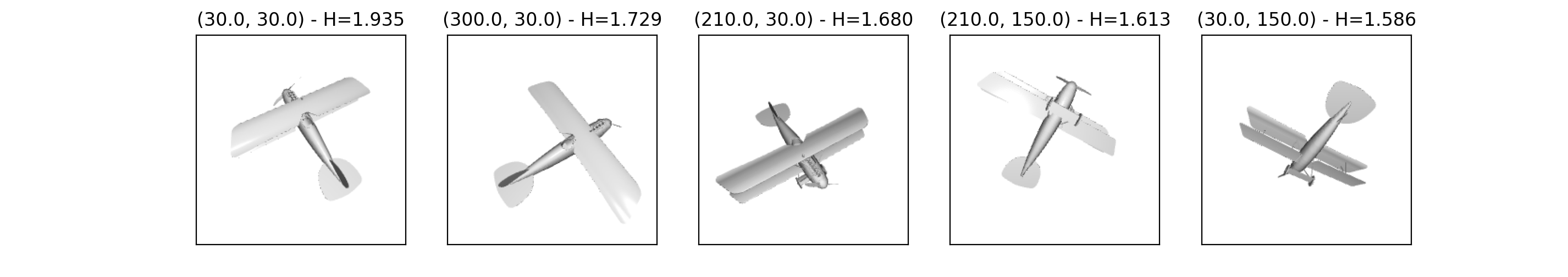}
    \includegraphics[width=\linewidth,trim={3cm 0.5cm 3cm 0}, clip]{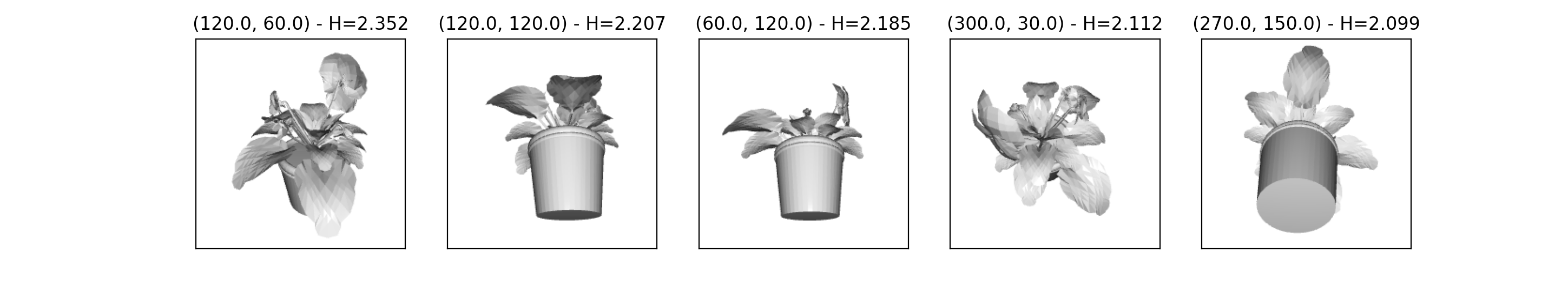}
    \vspace{-5mm}
    \caption{{Five best-views sorted by predicted entropy (more informative views) for unseen object classes: (\textit{top}) airplane,  and (\textit{bottom}) flower pot.}}
    \label{fig:unseentop5}
    \vspace{-4mm}
\end{figure}

{Higher predicted entropy values tend to denote more informative views. The figures 
{Fig}.~\ref{fig:unseentop5} show the five best-views of two never-seen-before objects (\textit{airplane} and \textit{flower pot}) predicted by the proposed model. For each of these views, the spherical coordinates are denoted by $\phi$ and $\theta$, while $H$ denotes the predicted entropy of these views. The obtained views are then used for classification and pose estimation purposes.} 

{It is also to be noted that the best view prediction model is independent of class information and the voxel grid does not need class labels. In the case of novel objects, temporary labels can be assigned based on the model's prediction confidence. Continual object recognition using MORE is out of the scope of this work and can be addressed by future research.}

\subsection{Multi-view classification and pose estimation}

As discussed in the introduction, we aim to design an approach to jointly handle object classification and pose estimation by learning shared high-level features. Following the notation from Elhoseiny et al. \cite{elhoseiny2016comparative}, our design falls in the category of Late Branching Models (LBM). As the backbone of {MORE}, we use an instance of a popular architecture, pre-trained on Imagenet~\cite{imagenet}, for image recognition, splitting the last layer into two fully connected layers with softmax activation of size $10$ and $60$ outputs for the object classification and the pose estimation respectively. Particularly, we evaluate MobileNetV2~\cite{sandler2018mobilenetv2}, and VGG-16~\cite{vgg}. The network was trained as a single-input multiple-outputs model. In particular, the model takes a single view as an input and it predicts the class of the originating object and the estimated viewpoint. The multi-view consists of the aggregation of $m$ single-view classifiers where $m$ is the number of views provided for the prediction. This method allows the network to accept a variable number of view images, to then return as outputs the classes represented by the majority votes.
While the object labels are quite straightforward to aggregate, different views result in different viewpoints. The predicted viewpoints are matched to the angle of the image views they were taken, the offset between these two values is the predicted rotation of the object from a standing front-facing position. Using these values, we can evaluate a majority vote for the pose estimation with precision up to half the distance between each of the $60$ originating viewpoints in the dataset ($15$ degrees on rotation around the z-axis and $15$ degrees on rotation around the y-axis). This precision can be utterly improved by generating a dataset with a more dense configuration of viewpoints and reshaping the network to classify a larger number of positions, with the cost of increasing the complexity of the network and the number of parameters. As regularization techniques, we used a dynamic learning rate. The learning rate set at the beginning of the training is $1e-4$, it then decreases on a plateau by a factor of 0.5 until reaching a minimum of $1e-8${.} This allows the learning process to switch to a progressively finer tuning in the later stages of the training. As loss functions, we used categorical cross-entropy for both class and pose prediction.

\begin{figure}[!t]
        
        \centering
        \includegraphics[width=0.8\linewidth, trim={0cm 0cm 0cm 0cm}, clip=true]{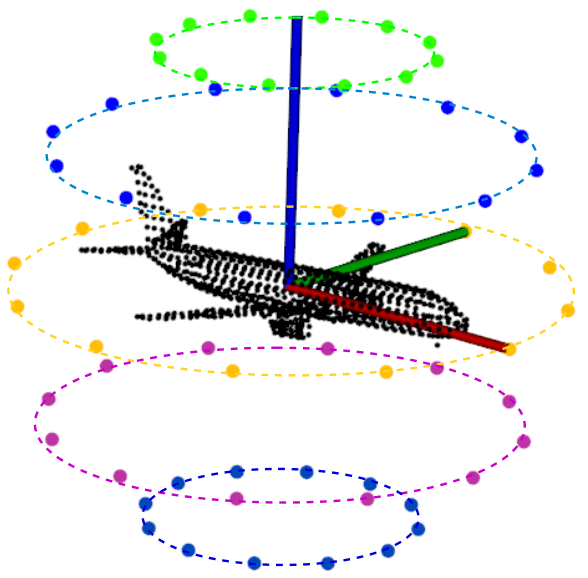}
        \vspace{-1mm}
        \caption{Viewpoint setup for rendering depth images for an airplane object: Colors indicate elevation levels, and distances between cameras and the center of the object are constant. The reference frame of the object is shown by red (x), green (y), and blue (z) lines.}
    \label{fig:viewsetup}
\end{figure}

\section{Results}
In this section, we first present the detail of generating a multi-view dataset for training {MORE} and then explain the evaluation metrics. Afterward, we discuss the performance of the proposed approach in the case of best-views prediction, single-view, and multi-view object classification and pose estimation. Finally, we integrated {MORE} into a robotic system and performed a set of real-robot experiments in the context of server fruit juice to show its usefulness in real-life scenarios.

\subsection{Dataset}
\label{dataset}
To build the dataset for the proposed model, we used the Princeton ModelNet40 dataset~\cite{wu20153d}, which consists of a collection of meshes from $40$ popular object categories. To generate depth images from a single 3D object model, virtual cameras are set up to point at the centroid of the object, then 2D depth images are rendered from each camera using a projection method. To achieve a consistent input we scaled each model to fit in a unit cube centered in the origin, we then subdivided the unit cube into a binary voxel grid, in which occupied voxels are shown by $1$ and the rest by zero. The obtained binary matrix represents the 3D model silhouette. We experimented with different grid sizes, while a higher number would have increased the resolution of the object, it would have increased exponentially the size of the data. We settled for a grid size of $50\times50\times50$ which offered an acceptable trade-off between size and resolution. The smoothing effect due to the little details in the object being cut off by the voxelization resolution happened to have a regularization effect in the learning since the convolution layers of the network would not try to learn such details as high-level features. To supply the closeness of the models to the sides of the cube due to the scaling and bounding process, we added a zero-padding of three voxels per side resulting in the occupancy grids to be of $56^3$ voxels.

\begin{figure}[!b]
    \centering
    \includegraphics[width=\linewidth, trim={0 0cm 0 0cm}, clip=true]{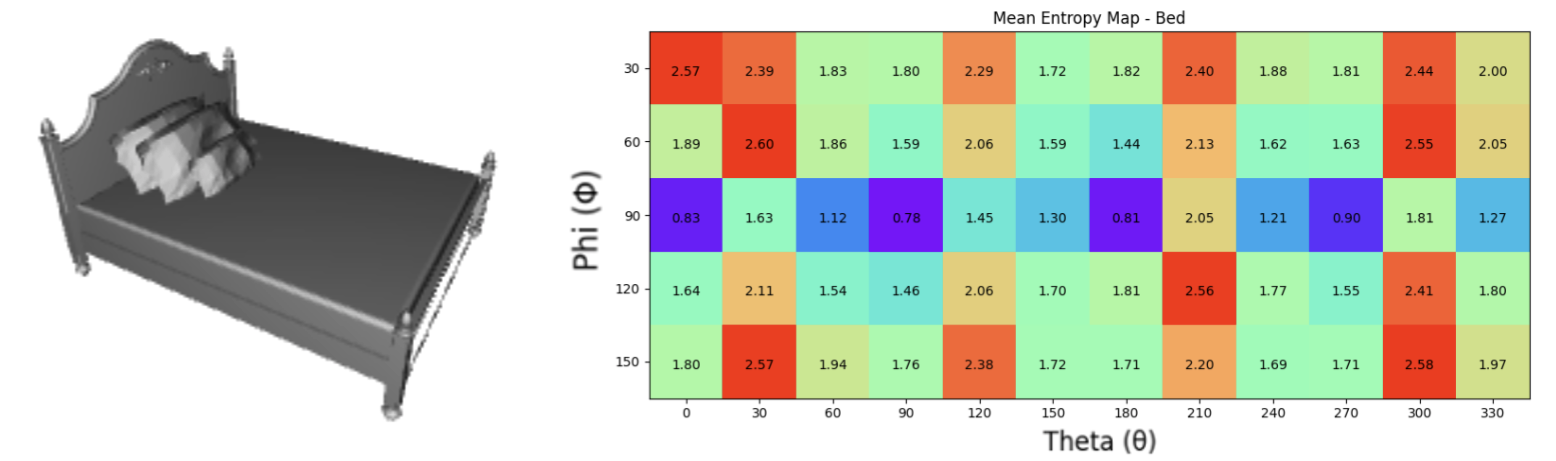}
    \caption{Average distribution of the $60$ views entropy for the class {bed}. The coordinates (x, y) of the graph indicate the rotation ($\theta$, $\phi$). $\theta$ represents the yaw angle, while $\phi$ represents the pitch angle. {The left image has been captured from ($\theta = 30 , \phi = 60$) coordinate.}}
    \label{fig:menmap2}
\end{figure}

We rendered a set of $60$ depth images from every object in the ModelNet40 dataset (see Fig.~\ref{fig:viewsetup}). We evaluated the quantity of information of every image with the Shannon's Entropy (see Eq. \ref{shannon}). 
To explain the idea better, we provide an example in Fig.~\ref{fig:menmap2}, representing the average entropy distribution for the category Bed. Given the shape of the average bed, the larger entropy values are found in correspondence to viewpoints closest to the four angles, since such viewpoints frame more faces of the cuboid.
It can be seen from the orange-red colours being predominant in the columns relative to the rotation around the z-axis at the values of $30$, $120$, $210$ and $300$ degrees. The lowest values instead are found on the row relative to the $90$ degrees rotation on the y-axis, meaning the object is being observed frontally. This is the worst angle to observe a cuboid since the upper and lower faces are not visible, hence resulting in the blue-violet row at $90$ degrees. We finally built the dataset by matching each 3D object to its entropy values. All of these processing steps were parallelized.

\subsection{Evaluation of best-views prediction model}
For input voxel grid of size $56\times56\times56$, we designed two branching $3$D convolutional layers with kernels sizes $3\times3\times3$ and $5\times5\times5$. The Hyperband algorithm \cite{li2017hyperband} evaluated $8$ as the best number of kernels for both the convolutional layers and $512$ units for the fully connected layer before the output layer. To accelerate and stabilize the learning process, we applied batch normalization in-between layers, and considered a progressively smaller learning rate (starting at $5\times10^{-5}$ until reaching $3\times10^{-7}$) and dropout factors of $0.25$ on the output of convolution layers and of $0.5$ on the fully connected layer. Since we formulate the problem in the form of a multi-output regression with $60$ values. To evaluate the quality of the learning, we used two measures: Mean Absolute Error (MAE) and Mean Squared Error (MSE), the former being employed as the loss function.

\begin{figure}[!t]
    \centering
    \includegraphics[width=\linewidth, trim = 1.5cm 3.7cm 1.5cm 0cm, clip=true]{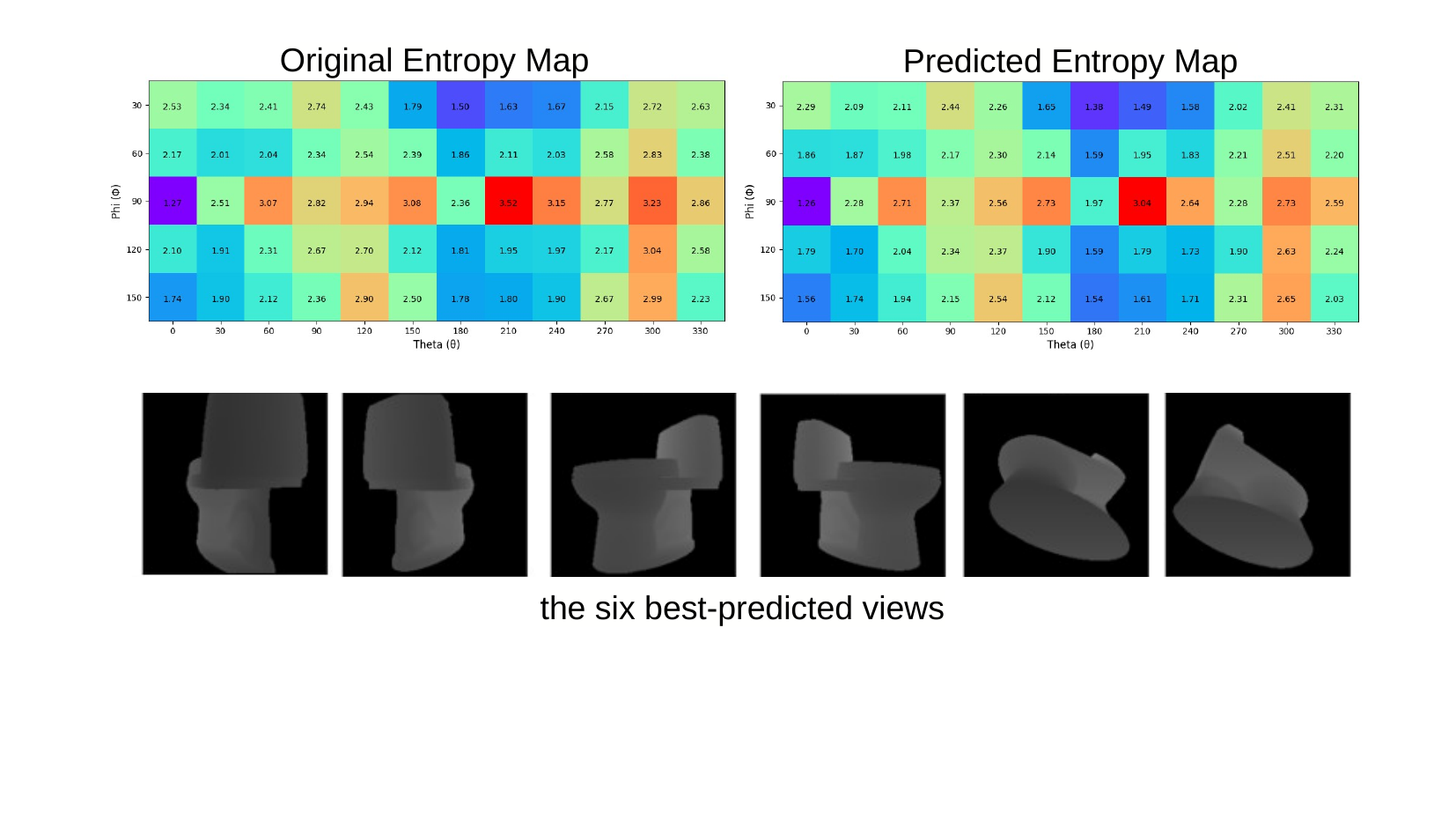}
    \caption{Entropy map calculated from the projected depth-views of a Toilet class object (\textit{top-left}) and the map of the same object predicted by {MORE} (\textit{top-right}). The colors indicate the entropy value calculated on a depth-image captured from the position ($\Theta,\Phi$). Violet-blue indicates smaller values while orange-red indicates larger values.  (\textit{lower-row}) The depth images for the six best-predicted views in order of entropy are shown on the lower row.}
    \label{fig:entropy-example}
\end{figure}
It was observed that the network was able to learn a function to approximate the entropy of $60$ viewpoints precisely. An example of a comparison between the original entropy map (built by calculating the entropy of views) and the map predicted by {MORE} for a Toilet object is shown in Fig.~\ref{fig:entropy-example}. The distribution of the predicted values resembles closely the distribution of the true values. 
To extract the best views from the entropy map we use a peak detection algorithm that returns the coordinates of the local maxima in the matrices.

\subsection{Evaluation of classification and pose estimation}

In this round of experiments, we first evaluated the proposed system using two core architectures, VGG-16~\cite{vgg} and MobileNetV2~\cite{sandler2018mobilenetv2}, which are reliable CNN architectures for object recognition. Both models are pre-trained using the ImageNet dataset~\cite{imagenet}, a large dataset consisting of $1.4$M images and $1000$ classes. The architecture is instantiated without the top layers, to adapt to the branching structure for object classification and pose estimation. The branching is performed at the last layer, following the Late Branching Model (LBM) example by the notation from Elhoseiny et al. \cite{elhoseiny2016comparative}. We experimented with different branching archetypes, however, the number of parameters increased dramatically without a corresponding improvement in accuracy. The last layer of the core architecture is split into two fully connected layers representing the number of possible classes and poses ($60$ nodes). We used the {Adam} optimizer with a starting learning rate of $1 \times 10^{-4}$ which is dynamically reduced on plateauing validation loss. As loss functions we employed {categorical cross-entropy} for both class and pose. 
We fine-tuned the architectures on a dataset of $293,940$ images, composed of projections from $60$ viewpoints of $4,899$ 3D models from the ModelNet10 dataset. For the ModelNet40 dataset, $8,617$ 3D models and $517,020$ images were used.

\subsubsection {Single view prediction}
\begin{figure}[!b]
    \centering
    \includegraphics[width=\linewidth, trim=0mm 0mm 0mm 0mm, clip=true]{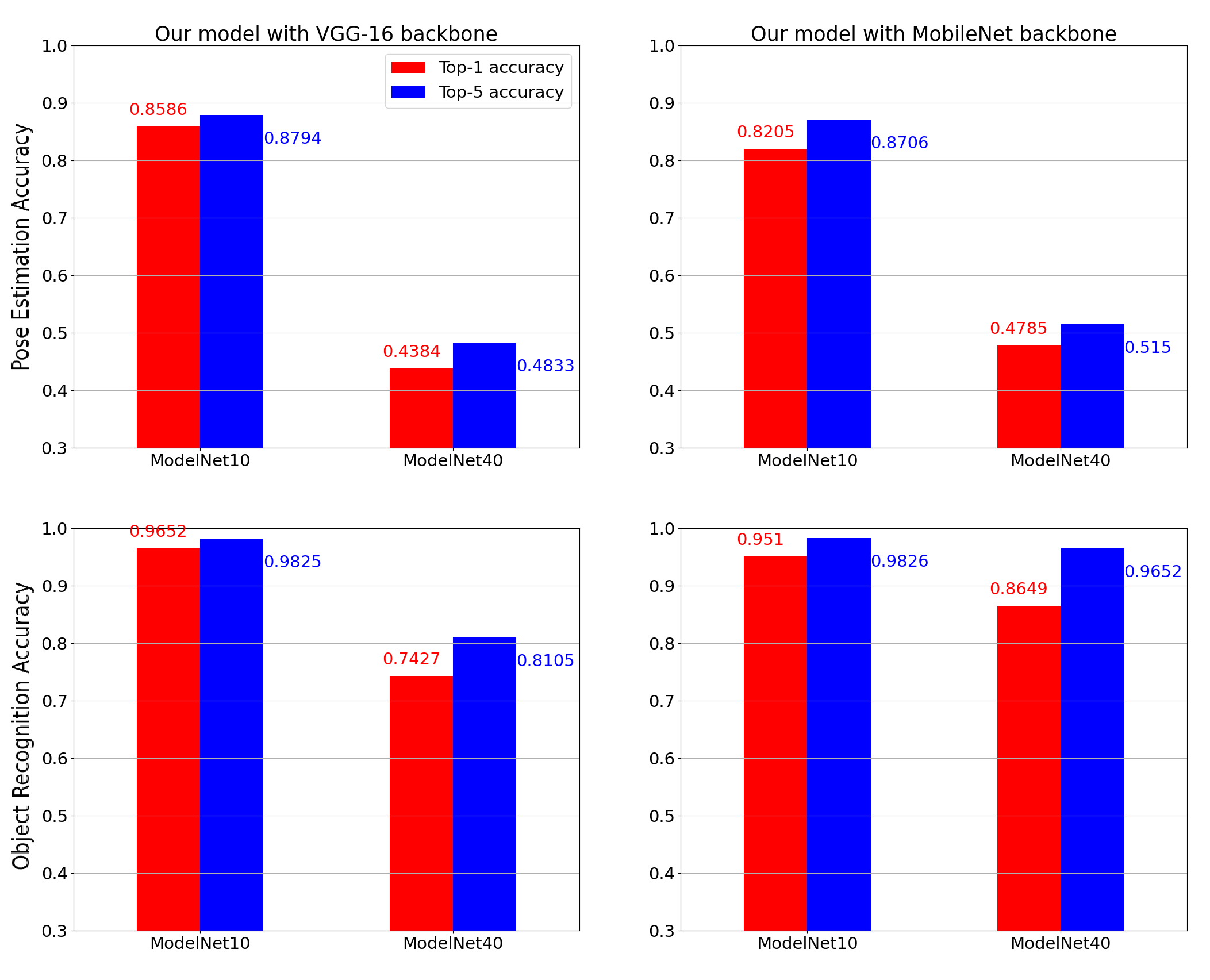}
    \caption{The performance of pose estimation (\textit{top-row}) and object recognition (\textit{lower-row}) of {MORE} on ModelNet$10$ and ModelNet$40$ datasets.}
    \label{fig:recognition_and_pose_estimation}
\end{figure}

{MORE} bases its final classification and pose estimation on a majority vote. Each prediction from the instances of the single-view CNN is pooled and contributes to the decision of the system (see the \textit{right-side} of Fig.~\ref{fig:overview}). We tested the proposed system with the best-performing models. For a fair comparison, we trained the single-view CNN and then tested the proposed approach using the training and test split of the original dataset as in \cite{wu20153d,kanezaki_rotationnet_2021}. We benchmark {MORE} against a few baselines, including LP-3DCNN \cite{kumawat2019lp}, iMHL \cite{zhang2018inductive}, RS-CNN \cite{liu2019relationshape}, Ma et al.\cite{ma_learning_2019}, Orthographic Net \cite{kasaei_orthographicnet_2020} and MLVCNN \cite{jiang2019mlvcnn}.  Results are summarized in Table \ref{tab:single-view-cnn} and Fig.~\ref{fig:recognition_and_pose_estimation}.

\begin{table}
	\centering
	\newcolumntype{?}{!{\vrule width 0.5pt}}
\setlength\arrayrulewidth{0.5pt}
	\caption{Performance of single-view object recognition.}
	\resizebox{0.7\columnwidth}{!}{%
			\begin{tabular}{|c|c|c|}
			\hline
			{Dataset} & Approach  & Acc. \footnotemark  \\ \hline
			\multirow{6}{*}{ModelNet10} & {MORE} -- MobileNet & \textbf{0.9826} \\\cline{2-3}
            & {MORE} -- VGG-16 & 0.9651 \\\cline{2-3}
            & LP-3DCNN~\cite{kumawat2019lp} & 0.9440 \\\cline{2-3}
            & Ma et al.~\cite{ma_learning_2019} & 0.9105 \\\cline{2-3}
            & Orthographic Net~\cite{kasaei_orthographicnet_2020} & 0.8856 \\\cline{2-3}
            & Primitive-GAN.~\cite{khan_unsupervised_2019} & 0.8640 \\
			\hline \hline
			\multirow{8}{*}{ModelNet40} & {MORE} -- MobileNet & \textit{0.9652} \\\cline{2-3}
            & {MORE} -- VGG-16 & 0.8105 \\\cline{2-3}
            & LP-3DCNN~\cite{kumawat2019lp} & 0.9210  \\\cline{2-3}
            & iMHL~\cite{zhang2018inductive} &  \textbf{0.9716}  \\\cline{2-3}
            & RS-CNN~\cite{liu2019relationshape} & 0.9360  \\\cline{2-3}
            & MLVCNN~\cite{jiang2019mlvcnn} &  0.9416 \\\cline{2-3}
            & Ma et al.~\cite{ma_learning_2019} & 0.9319\\\cline{2-3}
            & Primitive-GAN.~\cite{khan_unsupervised_2019} & 0.9220 \\
            \hline
		\end{tabular}
	}
	\label{tab:single-view-cnn}
\end{table}

By comparing the results obtained, it is visible that {MORE} with MobileNet backbone achieved the best recognition performance on ModelNet$10$ and the second best on ModelNet$40$. It also did much better than the VGG16 backbone. Experimental results showed that iMHL performed slightly better ($\approx 0.02$) than {MORE} concerning Object recognition on ModelNet$40$. The difference is most likely because we forced the network to learn a representation that can be used for object recognition and pose estimation. 

{In the case of the single view object recognition task, MORE performs better than other state-of-the-art models because it predicts the best view first and then recognizes the object. Unlike previous approaches, MORE finds views that contain rich information. This in turn contributes to better model performance.}

We plotted the confusion matrices to realize the differences between the prediction of VGG-16 and MobileNet architectures on ModelNet$10$ dataset.
It was observed that both architectures achieved more than $96\%$ accuracy over \textit{bed}, \textit{chair}, \textit{dresser}, \textit{monitor}, \textit{sofa}, and \textit{toilet} classes, and most of misclassifications mainly occurred within the \textit{bathtub}, \textit{desk}, and \textit{night-stand} categories. On closer inspection, we can see that the VGG-16 architecture misclassified \textit{bathtub} with \textit{sofa} more times than MobileNetV2, and also misclassified more frequently a \textit{desk} for other objects. While the overall accuracy on the \textit{desk} class is better for MobileNetV2, VGG-16 is more stable when separating the \textit{table} and \textit{desk} classes which are arguably the most difficult to distinguish. Another difference lies in the classification of the \textit{night-stand} class where VGG-16 performs significantly better than MobileNetV2. 
Overall both architectures seem to have difficulties in separating objects with very similar shape features. To solve this issue, backbone networks could be updated to encode both fine-grained and general features of input data simultaneously.


\subsubsection {Size of training data vs. performance}
Due to the large amount of data obtained, training the models using all data would be time consuming. We experimented with random subsets of the full dataset to check whether a fraction of it would achieve similar accuracy while reducing the training time required. The results of training models for both ModelNet$10$ and ModelNet$40$ are summarized in Table~\ref{tab:single-view-cnn-subsampled}. 
For each architecture, we tested the accuracy of both networks when trained on the full dataset, on a third ($1/3$) and a twentieth ($1/20$) of it.
\begin{table}[!t]
\centering
\caption{Single-view Object Recognition and Pose Estimation Accuracy.}
\setlength\arrayrulewidth{0.5pt}
\resizebox{\columnwidth}{!}{%
\begin{tabular}{|c|c|c|c|c|}
\cline{1-5}
\multicolumn{1}{|c|}{Dataset}  & \multicolumn{2}{c|}{{MORE} - MobileNet}     & \multicolumn{2}{c|}{{MORE} - VGG-16} \\ \cline{1-5}
\multicolumn{1}{|c|}{ModelNet10} & \multicolumn{1}{c|}{Class Acc.} & \multicolumn{1}{c|}{Pose Acc.} & \multicolumn{1}{c|}{Class Acc.} & Pose Acc.  \\ \cline{1-5}
\multicolumn{1}{|c|}{Full}                & \multicolumn{1}{c|}{0.9826}     & \multicolumn{1}{c|}{0.8706}    & \multicolumn{1}{c|}{0.9652}     & 0.8794     \\ \cline{1-5}
\multicolumn{1}{|c|}{1/3}                 & \multicolumn{1}{c|}{0.9713}     & \multicolumn{1}{c|}{0.8667}    & \multicolumn{1}{c|}{0.9618}     & 0.8600    \\ \cline{1-5}
\multicolumn{1}{|c|}{1/10}                & \multicolumn{1}{c|}{0.9697}     & \multicolumn{1}{c|}{0.8111}    & \multicolumn{1}{c|}{0.9558}     & 0.8611    \\ \cline{1-5}
\multicolumn{1}{|c|}{1/20}                & \multicolumn{1}{c|}{0.9496}     & \multicolumn{1}{c|}{0.8000}    & \multicolumn{1}{c|}{0.9119}     & 0.7444    \\ \hline\hline
ModelNet40 & Class Acc. & Pose Acc. & Class Acc. & Pose Acc. \\ \cline{1-5}
\multicolumn{1}{|c|}{Full}                & \multicolumn{1}{c|}{0.9652}     & \multicolumn{1}{c|}{0.5150}    & \multicolumn{1}{c|}{0.8105}     & 0.4833     \\ \cline{1-5}
\multicolumn{1}{|c|}{1/3}                 & \multicolumn{1}{c|}{0.9507}     & \multicolumn{1}{c|}{0.5009}    & \multicolumn{1}{c|}{0.7930}     & 0.4606    \\ \cline{1-5}
\multicolumn{1}{|c|}{1/10}                & \multicolumn{1}{c|}{0.9045}     & \multicolumn{1}{c|}{0.5000}    & \multicolumn{1}{c|}{0.7335}     & 0.4384    \\ \cline{1-5}
\multicolumn{1}{|c|}{1/20}                & \multicolumn{1}{c|}{0.8962}     & \multicolumn{1}{c|}{0.4640}    & \multicolumn{1}{c|}{0.7039}     & 0.4320    \\ \cline{1-5}
\end{tabular}}
\vspace{-3mm}
\label{tab:single-view-cnn-subsampled}
\end{table}

The accuracy metrics describe the ratio of labels, on the test split of the dataset, that are correctly predicted by the model from a single view image. We consider the pose to be correctly predicted when the model reports the coordinates $(\Theta, \Phi)$ corresponding to the viewpoint used to capture the view. We do see that increasing the amount of training data does lead to an increase in performance, but for the ($1/2$) and ($1/3$) this does not make a drastic difference. Understandably, for ($1/20$), there is a huge hit to the performance. Even with this reduction though, the model performs well which shows the validity of {MORE}.

\begin{table}[!t]
\newcolumntype{?}{!{\vrule width 0.5pt}}
\setlength\arrayrulewidth{0.5pt}
    \centering
    \caption{Accuracy of Multi-view Object Recognition and Pose Estimation On ModelNet10.}
    \begin{tabular}{c c c}
    \hline
     Model & Class Accuracy & Pose Accuracy \\
     \hline
     VGG-16         & $0.9020$ & \textbf{$0.9394$} \\
     MobileNetV2 & \textbf{$0.9130$} & $0.9372$ \\
     \hline
    \end{tabular}
    \label{tab:mv_scores}
\end{table}
\subsubsection{Multi-view prediction}
\begin{figure}[!t]    
    \centering
    \includegraphics[width=\linewidth, trim={0 4mm 0 0cm}, clip=true]{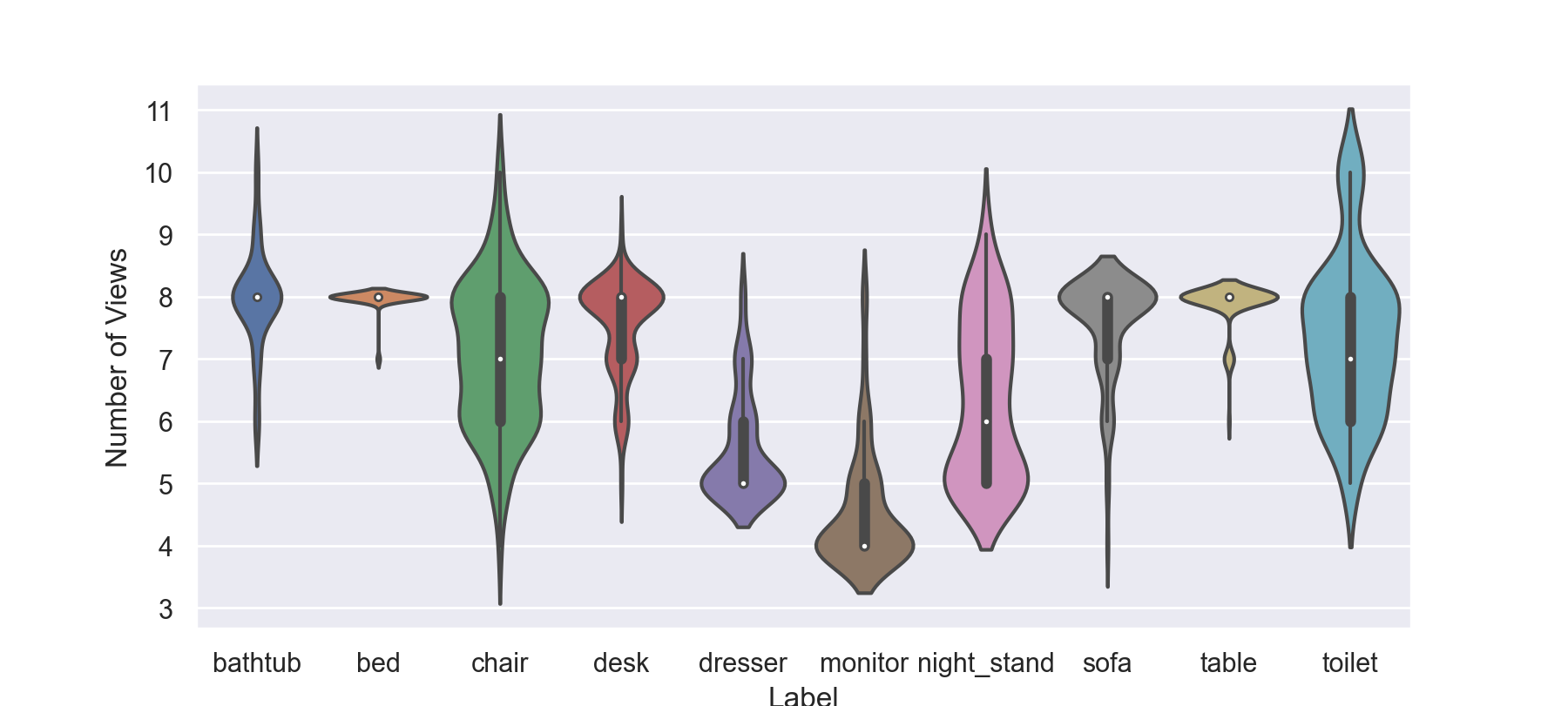}
    \caption{Violin plot of the distribution of the number of selected best-views for each category from the ModelNet10 dataset.}
    \label{fig:violinplot}
\end{figure}

As for the views used by the model to make the predictions, the peak detection algorithm makes the best-view selection and outputs several views equal to the number of local maxima in the entropy map. Results are reported in Table~\ref{tab:mv_scores}. By comparing the results, it is clear that the model with MobileNet achieved better recognition accuracy than the VGG-16, while the VGG-16 achieved better pose estimation accuracy. An overview of the distribution of the number of views used for each category is presented in Fig.~\ref{fig:violinplot}. As it is shown in the graph the model adapts the number of views it uses for the prediction to the supposed complexity of the object it is observing. On average the algorithm selects seven views to perform the prediction. 
The mean of inference time for $10$ simultaneous classification and prediction for the single-view method on an NVIDIA RTX2070 is $0.048\pm 0.013$ seconds.  


\begin{figure*}[!t]
    \centering
    \includegraphics[width=\linewidth, trim={0  00mm 0 0cm}, clip=true ]{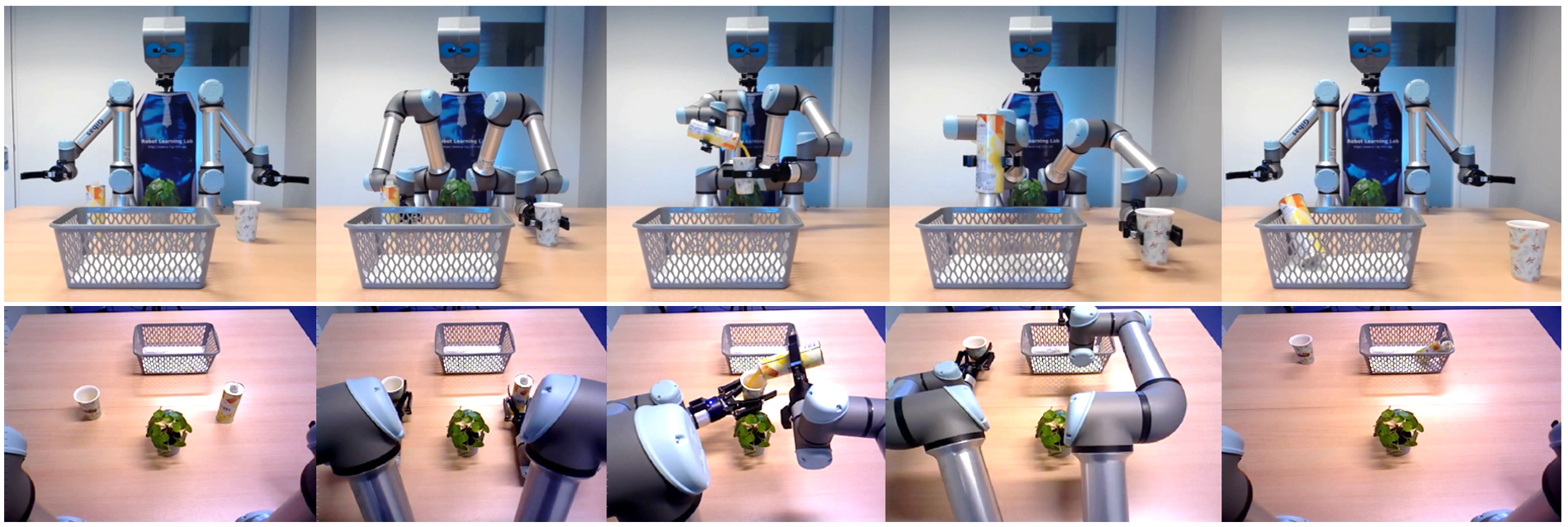}
    \caption{A series of snapshots demonstrating the performance of our dual-arm robot in a scenario where it serves orange juice: to accomplish this task the robot should detect the pose and label of all objects, and then grasp and manipulate the juice bottle and the mug into the serve position. The juice is then poured into the cup by the robot. Finally, the robot hands the user the cup and places the juice bottle in the basket. (\textit{top-row}) images captured from an external camera; (\textit{lower-row}) images captured from the robot's camera.}
    \label{fig:real_robot_seq}
\end{figure*}

\subsection {Robustness}

\begin{figure}[!b]
    \centering
    \includegraphics[width=\linewidth, trim={0  00mm 0 0cm}, clip=true]{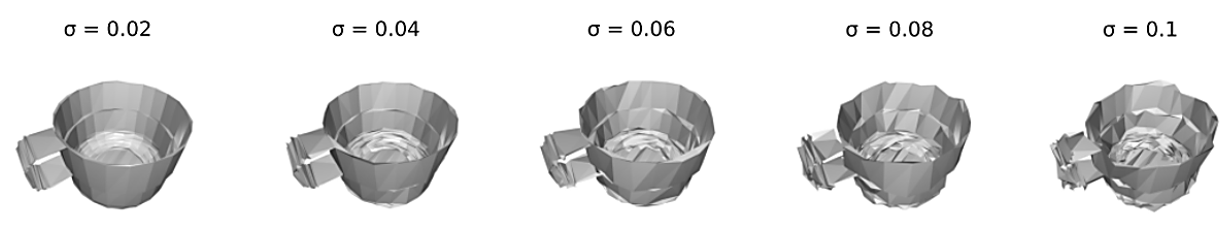}
    \caption{An illustration of a cup object with five levels of Gaussian noise (i.e., various $\sigma$ value). We applied noise {to the vertices of the mesh file of the object} in all three axes of the object.}
    \label{fig:sigmax}
    \vspace{-2mm}
\end{figure}

\begin{wraptable}{!t}{0.5\linewidth}
\vspace{-2mm}
\centering
\newcolumntype{?}{!{\vrule width 0.5pt}}
\setlength\arrayrulewidth{0.5pt}
\caption{Effect of Gaussian Noise on Object recognition accuracy on ModelNet40}
\label{tab:def}
\resizebox{\linewidth}{!}{%
\begin{tabular}{|c|c|c|}
\hline
$\sigma$    & VGG16 & MobileNetV2 \\ \hline
0.02 & 0.8303           & 0.9212                 \\ \hline
0.04 & 0.7333           & 0.7333                 \\ \hline
0.06 & 0.6606           & 0.6484                 \\ \hline
0.08 & 0.5333           & 0.5818                 \\ \hline
0.10 & 0.4060           & 0.5333                 \\ \hline
\end{tabular}%
}
\end{wraptable}
We conducted another series of experiments to determine how robust the proposed approach is to Gaussian noise, circumstances that frequently occur in real-life settings. In particular, we tested the performance of {MORE} on varying standard deviation ($\sigma$) values for Gaussian noise. {We applied noise to the vertices of the mesh file of the object to mimic the effects of information loss while viewing the object.} An example of this noise with varying ($\sigma$) values is shown in Fig.~\ref{fig:sigmax}.

For these experiments, we used the ModelNet40 dataset as it would be more representative of the robustness of our solution as it has more classes and is more challenging compared to the ModelNet10. Results are reported in Table~\ref{tab:def}. We can see that the model did well for a majority of the cases even with this noise. Having a very high noise led to a drastic loss in the accuracy of the prediction which is well within expectation. It is to be noted that even with a slight noise, there was a slight drop in performance. The MobileNet model did better in this experiment as well, compared to the VGG-16.

\subsection{Real-robot demonstration}

To show all the functionalities of the proposed approach, we integrated it into a cognitive robotic system. Our robotic setup consists of two UR5e arms, which are equipped with Robotiq 2F-140 gripper, and an Asus Xtion RGB-D camera as shown in Fig.~\ref{fig:real_robot_seq}. In this round of the experiment, we fine-tuned {MORE} using real point cloud data.  Initially, we randomly place a cup, a flower, a bottle of juice, and a basket in front of the robot, and then instructed the robot to perform \textit{Serve\_Fruit\_Juice} task.

\begin{figure}[!t]
    \centering
    \includegraphics[width=\linewidth, trim={0  00mm 0 0cm}, clip=true]{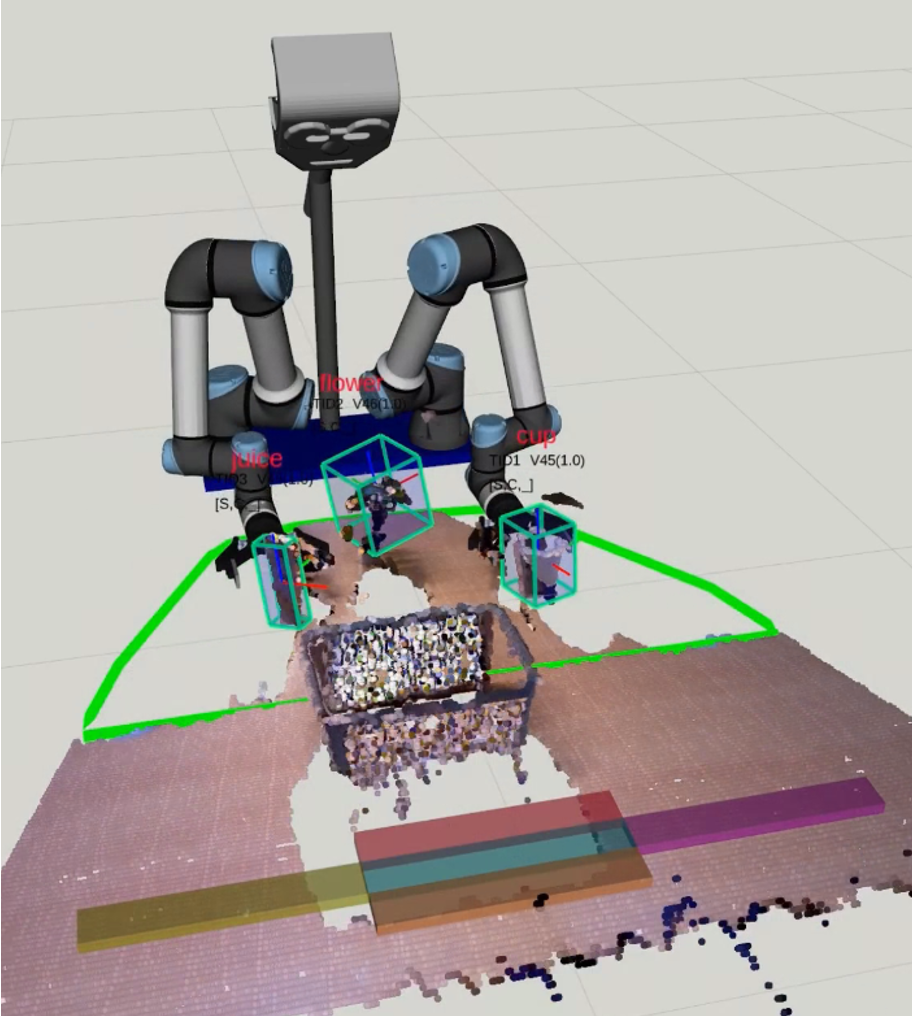}
    \caption{Visualizing the perception of the robot in Rviz during the ``\textit{serve\_fruit\_juice}'' scenario: The robot’s workspace is shown by the green convex-hull. The pose of each object is shown by the bounding box and its reference frame. The recognition results are visualized above each object.}
    \label{fig:rviz}
    \vspace{-2mm}
\end{figure}

To accomplish this task, the robot needs to recognize all the objects and estimate their poses accurately. It should be noted that while the label and the pose of the other objects (\textit{cup}, \textit{juice}, and \textit{flower}) should be detected, the pose of the basket is known in advance. Toward this goal, we first removed the points belonging to the dominant plane (table) using the RANSAC algorithm and then applied euclidean clustering to the remaining points. We considered each of the clusters as an object candidate and fed them into our pipeline to estimate their pose and label.  
After recognizing all the objects and estimating their pose, the robot should plan collision-free trajectories for the left and right arms to grasp the bottle of juice and the cup (see Fig.~\ref{fig:rviz}). After grasping the objects, the robot manipulates them into the severing position and then pours the juice into the cup. Finally, the robot delivers the drink to the user and places the bottle into the basket, and returns to the initial pose. A sequence of snapshots demonstrating the performance of the robot is shown in Fig.~\ref{fig:real_robot_seq}.
We repeated these experiments five times. In all experiments, the robot was able to serve the drink and place the juice bottle into the basket successfully. 

\begin{figure*}
    \centering
    \includegraphics[width=1\linewidth, trim={0 0cm 0 0cm}, clip]{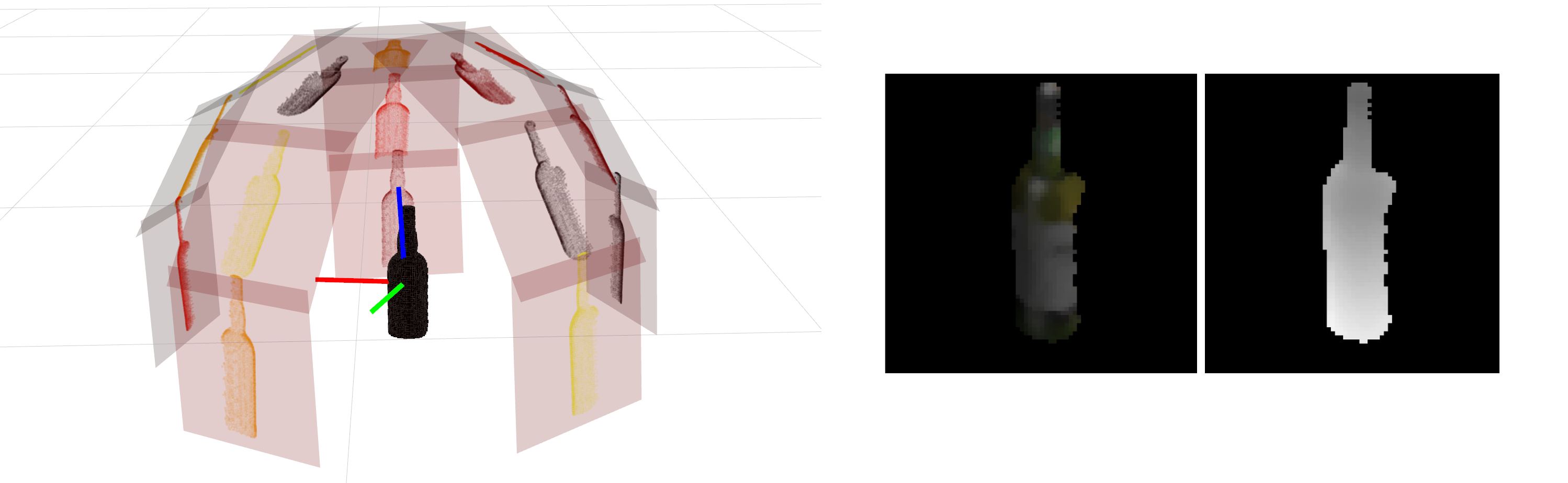}\\
    \includegraphics[width=1\linewidth, trim={0 0cm 0 0cm}, clip]{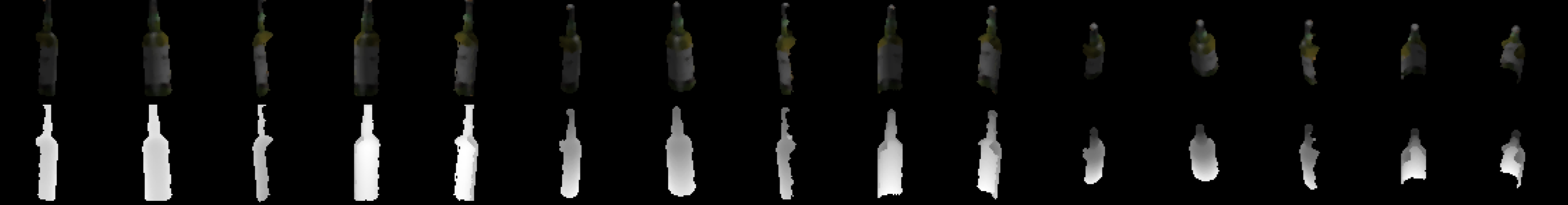}
    
    \caption{{An example of generating 15 views of a bottle object using a single RGB-D camera based on an orthographic projection technique \cite{kasaei2023mvgrasp}: (\textit{top-left}) point cloud of the object, its reference frame, and the projection of the object on 15 plans uniformly distributed over a hemisphere. The red, green, and blue lines show the X, Y, and Z axes respectively. (\textit{top-right}) One of the selected RGB and depth views of the object. (\textit{lower-row}) Multiple RGB-D views of the object are generated from 15 different perspectives distributed over a hemisphere around the object. Note that, the object is partially visible in some of the images due to self-occlusion.}}
    \label{fig:mvgen}
    \vspace{-5mm}
\end{figure*}

{In the case of real-world scenarios where it is not possible to change the camera pose or use multiple cameras, an orthographic projection technique \cite{kasaei2023mvgrasp} can be used to generate multiple views of the target object based on the information that can be seen from a single RGB-D camera. To discuss this point better, we captured a partial point cloud of a bottle object using an Asus Xtion RGB-D sensor and then used an orthographic projection technique~\cite{kasaei2023mvgrasp} to generate 15 RGB-D views of the bottle from various perspectives (see Figure~\ref{fig:mvgen} (\textit{top-row})). In this example, as shown in Figure~\ref{fig:mvgen} (\textit{top-left}), views are uniformly distributed over a hemisphere and visualized by different colors. The generated RGB-D views of the object are shown in  Figure~\ref{fig:mvgen} (\textit{lower-row}). This technique allows us to estimate the images of the target object from different perspectives while using only one camera sensor.}
{Furthermore, we hypothesize that better performance can be obtained by applying shape completion techniques to the point of the object \cite{varley2017shape}. Alternatively, we can generate a completed point cloud of the scene by increasing the number of available cameras and using point cloud registration techniques \cite{xu2019rgb,wong2017segicp}}.

\section{Conclusions}
In this paper, we proposed a deep learning-based approach to tackle the simultaneous recognition and pose estimation of 3D objects. We suggested a deep object-agnostic entropy estimation model, capable of predicting the best viewpoints of a given object. We then used the obtained views of the object to predict the pose and category label of the target object simultaneously. Experimental results showed that the predicted views of objects are descriptive enough to achieve high accuracy scores in both classification and pose estimation tasks. To show the usefulness of the {MORE} in real-life scenarios, we integrated it into a robotic system and performed the serve fruit juice task with a dual arm robot. In continuation of this work, 
we would like to investigate fine-grained pose estimation and object recognition. 
Another potential avenue to look into is to scale up the training set using synthetic data. In particular, the discrete nature of the proposed pose estimation leaves the sensitivity of the pose estimator depending on the density of the dataset, hence it would be possible to achieve more precise estimations by sampling a dataset with a larger number of viewpoints.

\section*{Author Contribution}
\noindent
Hamidreza Kasaei proposed the main idea and designed the work. Tommaso Parisotto and Subhaditya Mukherjee developed the idea and performed experiments. Hamidreza Kasaei supervised the study, developed part of the code, and helped with real robot experiments. All authors provided critical feedback and helped shape the research and manuscript. {Note that this paper is an extension of the Master’s thesis of Tommaso Parisotto, which was done under the supervision of Hamidreza Kasaei.}

\section*{Ethical Statement}
\noindent
Each of the authors confirms that this manuscript is their own original work and has not been previously published elsewhere and is not currently under consideration by any other journals. Additionally, all of the authors have approved the contents of the paper and have agreed to the submission policies of Autonomous Robots.

\section*{Acknowledgements} 
\noindent We thank the Center for Information Technology of the University of Groningen for their support and for providing access to the Peregrine high-performance computing cluster.

\section*{Founding} 
\noindent 
There is no funding to report for this submission.
\bibliographystyle{IEEEtran}
\bibliography{reference}

\end{document}